%% file: main.tex
\documentclass{article}

\PassOptionsToPackage{sort,numbers,compress}{natbib} %
\usepackage[algo2e,ruled,vlined,linesnumbered]{algorithm2e}
\usepackage[dblblindworkshop, final]{neurips_2026}
\workshoptitle{Foundations of Agentic Systems Theory (FAST)}

\usepackage[utf8]{inputenc}
\usepackage[T1]{fontenc}
\usepackage{hyperref}
\usepackage{url}
\usepackage{booktabs}
\usepackage{graphicx}
\usepackage{titletoc}
\usepackage{amsmath,amssymb,amsthm}
\usepackage{nicefrac}
\usepackage{microtype}
\usepackage{enumitem}
\usepackage{xfrac}
\usepackage{xspace}
\usepackage{soul}
\usepackage{subcaption}
\usepackage[dvipsnames,table]{xcolor}
\usepackage{wrapfig}

\SetKwInOut{Require}{Require}
\usepackage[capitalize,noabbrev]{cleveref}
\usepackage{makecell}
\usepackage{siunitx}
\usepackage{multicol}
\usepackage{multirow}
\usepackage{tikz}
\usepackage{mathrsfs}
\usepackage{setspace}
\usepackage{listings}
\usepackage{bbm}
\usepackage{float}

\definecolor{llamacolor}{HTML}{eda100}
\definecolor{mistralcolor}{HTML}{1baf7a}
\definecolor{granitecolor}{HTML}{2a78d6}
\definecolor{qwencolor}{HTML}{e34948}

\definecolor{aliceblue}{HTML}{F0F8FF}
\definecolor{lavenderblush}{HTML}{FFF0F5}
\definecolor{oldlace}{HTML}{FDF5E6}
\definecolor{mintcream}{rgb}{0.96, 1.0, 0.98}
\definecolor{lightred}{RGB}{255, 230, 230}
\definecolor{oppvar}{HTML}{9B4A4A}

\newcommand{\cmark}{{\color{green!55!black}\ensuremath{\checkmark}}}
\newcommand{\xmark}{{\color{red!70!black}\ensuremath{\times}}}

\newsavebox{\chartbox}
\newsavebox{\tablebox}

\theoremstyle{definition}

\newcommand{\game}{\mathcal{G}}

\newcommand{\Xcal}{\mathcal{X}}
\newcommand{\simplex}{\Delta}
\newcommand{\gain}{g}
\newcommand{\cost}{c}
\newcommand{\val}{v}
\newcommand{\ctx}{x}

\title{Choosing LLMs for Strategic Agent Interactions}

\author{%
  Arastun Mammadli \\
  University of Exeter \\
  United Kingdom \\
  \texttt{am2169@exeter.ac.uk} \\
  \And
  Herbert Woisetschläger \\
  IBM Research \\
  Switzerland \\
  \texttt{h.woisetschlaeger@ibm.com} \\
  \And
  Jawad Fayaz \\
  University of Exeter \\
  United Kingdom \\
  \texttt{j.fayaz@exeter.ac.uk} \\
  \And
  Shiqiang Wang \\
  University of Exeter \\
  United Kingdom \\
  \texttt{s.wang9@exeter.ac.uk} \\
}

\begin{document}

\maketitle

\begin{abstract}
Game-theoretic benchmarks claim that large language model (LLM) strategic decision-making varies across game types, interaction structures, and capability dimensions.
However, they only evaluate strategic performance across individual fixed models.
In parallel, contemporary LLM routers pick models under quality-cost trade-offs, but with no notion of a strategic structure.
They do not account for an opponent that might be unknown, and that can adapt over time and shift the feedback signal.
To address this, we extend the game-theoretic evaluations and define strategic selection as the meta game layered on top of the base game.
In each round, a strategic selector fields an LLM from a portfolio of models to navigate a game payoff matrix that is latent and learned online.
We find that the evaluation signal is game-specific, and that extracting a reliable signal requires a much higher number of seed runs $N$ compared to traditional benchmarking setups.
Evaluating across 5 games and 10 LLMs, we show a model's strategic competence varies significantly depending on the game setup and the opponent faced.
Building on this, we implement a suite of selector methods and find that approaches that rely on query embeddings or LLM self-reflection do not consistently capture the underlying strategic signal.
A lightweight multi-armed bandit algorithm performs more reliably across games, though it still falls short of the best fixed model.
Finally, we show that simple non-stationary and adversarial bandit variants fail to outperform a stationary bandit baseline, pointing to a gap in how current methods handle a shifting opponent in strategic settings.
\end{abstract}

\begin{figure}[H]
  \centering
  \includegraphics[width=0.95\textwidth]{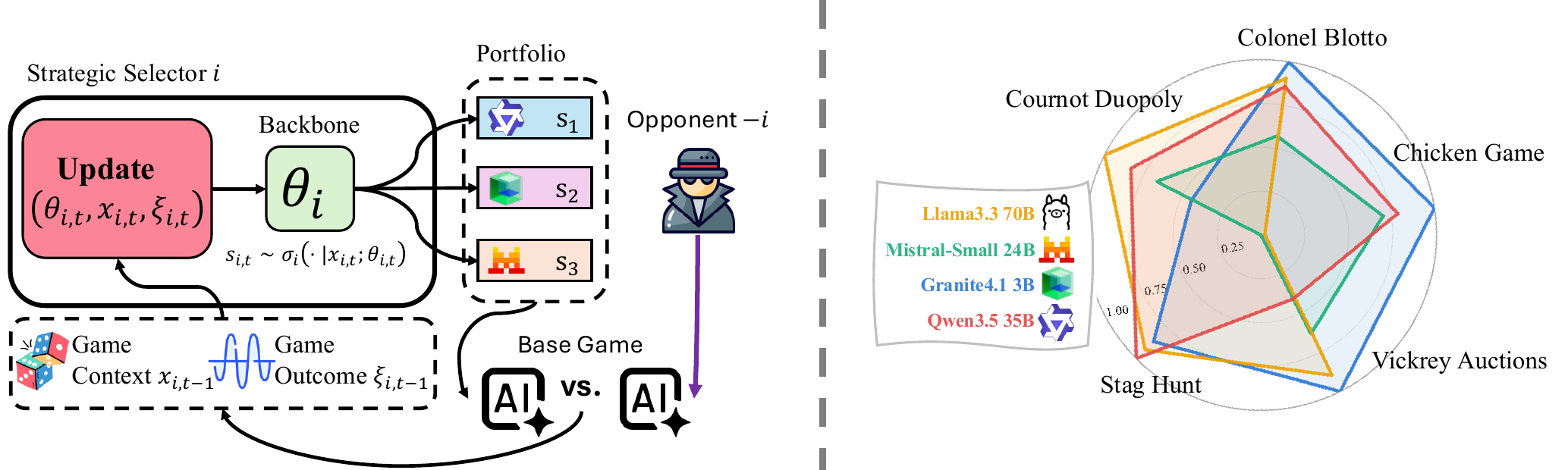} 
  \caption{\textit{Left:} A strategic selector $i$ plays a repeated game over $T$ rounds against an unknown opponent $-i$ and observes the game outcome $\xi_{i, t}$.
  The $\textsc{Update}(\theta_{i, t}, \ctx_{i, t}, \xi_{i, t})$ function is applied at every round $t$.
  \textit{Right:} Strategic performance (game utility $\val$) varies across games and models.\vspace{1em}}
  \label{fig:motivation}
\end{figure}

\section{Introduction}
\label{sec:intro}

Large language models (LLMs) are increasingly deployed in multi-agent settings as part of a larger compound system.
However, a growing share of deployers treat the inter-agent dynamics, competitive and cooperative alike, as a black box.
This makes it hard to build systems that remain capable, efficient, and reliable.
Existing work addresses pieces of this problem in isolation.
The strategic decision-making literature focuses on a fixed model's play under a specified game setting \citep{GTBench, TMGBench, Marshal}.
In parallel, LLM routing addresses model selection under performance-cost trade-off \citep{RouterDC, CausalLLMRouting, SLARouter}, but with a static task distribution and no notion of a strategic structure or opponent.
Neither accounts for the fact that contemporary LLM portfolios differ not only in scale and cost, but also in the type and quality of strategic behaviour they exhibit.

Our work reframes model selection itself as a strategic decision.
Concretely, we extend the game-theoretic evaluations and define strategic selection as the opponent-conditioned meta game layered on top of the base game.
Here the base game is played between the fielded LLMs, while the meta game is played between two strategic selectors that field a model of their choice each round.
In repeated settings, the dynamic and adaptive nature of this problem (as opposed to traditional routers operating on a fixed distribution of tasks) means our selection should be strategic.
Beyond balancing the performance-cost trade-offs, such a selection layer can help steer the deployer's general play, adding control over the reliability and safety of the interaction (\Cref{fig:motivation}).
We ask the question:
\begin{quote}
\itshape
Given LLM strategic capabilities vary across different games and models, can a
selector exploit this by learning which model to field against an unknown and possibly adaptive opponent?
\end{quote}

\textbf{Why This Matters.}
Traditional routers operate under the assumption of a fixed task distribution, learning a static mapping from queries to optimal models. 
In contrast, operating against an adaptive opponent transforms this into a non-stationary learning problem, where the optimal strategy can continuously shift over time.
A further challenge arises from the partial observability of the meta game's payoff matrix. 
In practice, a selector cannot observe the opponent's net gain, since the opponent's model choice is unknown, nor can it observe the counterfactual payoffs of fielding an alternative model in the current game state.
We also acknowledge that LLMs are stochastic models.
In strategic selection, we introduce an additional layer of environmental noise through the dynamic responses of the opponent.
Therefore, benchmarking and result reporting call for a careful statistical treatment.
Our contributions are threefold:
\begin{itemize}[noitemsep, topsep=2pt, leftmargin=*]
\item \textbf{Strategic Model Selection.}
  We formalise strategic selection as a repeated general-sum meta game between two players.
  Beyond capability and cost, this lets us treat the interaction itself as a first-class design consideration for model selection.
\item \textbf{Quantifying Noise in Strategic Play.}
  We characterise the signal and noise levels across games, and suggest adapting methods from traditional hypothesis testing to decide on the number of seeds per a game-portfolio pair.
\item \textbf{Empirical Investigation.}
  To set up our empirical investigation, we select 10 models from different families, and adapt a range of existing routing methods (\Cref{tab:selectors}).
  We first show that LLM strategic capabilities are scenario-dependent and exploitable by a selector.
  We then investigate the performance of adapted methods against both fixed models and selector opponents.
\end{itemize}

\begin{table}[t]
\centering
\caption{Implemented suite of selector methods. 
\textit{Cost-Aware}: the method accounts for a quality-cost trade-off when selecting a model. 
\textit{Opp.-ID}: the opponent's identity is revealed to the selector. 
\textit{Online}: the method learns online (round-by-round) from game outcomes.
\textit{Embed.}: the method encodes context into embeddings before feeding them into the predictor.
\textit{Non-Stat.}: the method is designed to handle a non-stationary environment.}
\label{tab:selectors}
\footnotesize
\setlength{\tabcolsep}{3pt}
\renewcommand{\arraystretch}{0.9}
\begin{tabular}{@{}lcccccl@{}}
\toprule
\textbf{Method} & \textbf{Cost-Aware} & \textbf{Opp.-ID} & \textbf{Online} & \textbf{Embed.} & \textbf{Non-Stat.} & \textbf{Source} \\
\midrule
Fixed Model                     & \xmark & \xmark & \xmark & \xmark & \xmark & --- \\
Per-Opponent Full-Information   & \cmark & \cmark & \xmark & \xmark & \xmark & --- \\
\midrule
\multicolumn{7}{@{}l}{\textit{Implemented selectors}} \\
LLM Selector                     & \cmark & \xmark & \cmark & \xmark & \xmark & \citet{ReflexionFramework} \\
RouterDC                        & \cmark & \xmark & \xmark & \cmark & \xmark & \citet{RouterDC} \\
Causal LLM Router              & \cmark & \xmark & \cmark & \cmark & \xmark & \citet{CausalLLMRouting} \\
Thompson Sampler                & \cmark & \xmark & \cmark & \xmark & \xmark & \citet{ThompsonSampler} \\
Discounted Thompson Sampler     & \cmark & \xmark & \cmark & \xmark & \cmark & \citet{NonStationaryBandit} \\
Sliding-Window Thompson Sampler & \cmark & \xmark & \cmark & \xmark & \cmark & \citet{SlidingWindowBandit} \\
EXP3                            & \cmark & \xmark & \cmark & \xmark & \cmark & \citet{EXP3Paper} \\
\bottomrule
\end{tabular}
\end{table}

\section{Related Work}
\label{sec:related-work}

\subsection{Strategic LLM Decision-Making}
A growing literature explores LLMs in strategic settings (e.g. how LLMs play games against each other).
Game-theoretic benchmarks \citep{GTBench, TMGBench, GAMA-Bench, VSBench} and frameworks \citep{ECON, BehavioralGT, Marshal, Memo} test LLMs in normal-form games and find that performance is uneven and structure-dependent.
Others \citep{Smartplay, Magic, DSGBench, LLMArena} evaluate capability on a higher level (e.g. planning, collaboration, theory of mind) and consistently find spiky, uneven profiles across models.
Similarly, \citet{TuringTest} and \citet{HomoSilicus} run behavioural Turing tests and show that different models exhibit different strategic dispositions (i.e. behavioural profiles).
Motivated by such findings, \citet{STEER} outlines the need for an element-level (model, prompt, domain) selection approach, but leaves it to future work.
We note that despite extensive coverage, the existing literature always fixes the model per player.

\subsection{LLM Routing}
An orthogonal direction learns to predict a best-fit model for a given query, often trading off performance for cost and relying on minimal two-model setups \citep{RouteLLM, HybridLLM, AutoMix}.
Others extend this to incorporate more models \citep{FrugalGPT, xRouter, CARROT, MESS+} or train on limited observational data \citep{CausalLLMRouting, SLARouter}.
These works represent queries as vector embeddings to semantically capture their scope and difficulty.
However, we argue that discriminative signal in strategic selection largely comes from evolving game history (actions and outcomes), and two consecutive rounds with near identical embeddings can call for opposite model choices.
In the absence of routing methods that explicitly model this structure, we also consider bandit-style per-round feedback \citep{ThompsonSampler, EXP3Paper} as a paradigm suited to a non-stationary interactive signal.

\section{Strategic LLM Selection}
\label{sec:formulation}

\subsection{What is Strategic Selection?} 
In multi-agent systems, a setting is \textit{non-strategic} if the objective concerns only the overall performance of the system or its performance-cost trade-off.
A setting is \textit{strategic} if it also accounts for the interaction between agents, including how they communicate, coordinate, and negotiate.
Modelling this interaction can produce systems that are more reliable and efficient.
In a cooperative setting, it improves the coordination between agents \citep{CooperBench}.
In a competitive setting, it gives the deployer more control over the interaction with an opponent and helps steer the interaction toward the deployer’s preferences.
For example, for a given business, the reward of the negotiation, the cost of using AI, and the consistency (i.e. associated risk) of the results can all matter at varying levels.

\subsection{Problem Formulation}
We formalise strategic LLM selection as a repeated, contextual, general-sum game with two layers: a \textit{base game} that fielded models play, and a selection \textit{meta game} that the deployers play.
Strategic selection focuses on the meta layer, a two-player game with $i$ indexing the player (deployer) and $-i$ its opponent.
Repeated games are played over rounds $t = 1, \dots, T$, where each round $t$ is associated with context $x_{i, t} \in \Xcal$ (e.g. game setup, instructions, past history).
Deployer $i$'s set of pure strategies (meta game actions) $S_i$ is defined as the set of models $\{1,\dots,M_i\}$ in its portfolio.
In each round, $i$ selects a model $s_{i,t}\sim\sigma_i(\cdot\mid\ctx_{i, t})$, where $\sigma_i\in\simplex(S_i)$ is a mixed strategy (i.e. selection policy).
Per-round context $\ctx_{i, t}$ is then passed down to the fielded model to produce the base game action $s_{i, t}(x_{i, t}) \to a_{i,t}$.
The base game (i.e. of fielded models $s_{i, t}$ against $s_{-i, t}$) produces a utility $\val_{i,t}$ at an inference cost $\cost_{i,t}$ (tokens spent), as well as a net gain of $\gain_{i,t}$ that captures their trade-off.
Together, player $i$'s game outcome is denoted as $\xi_{i, t} = (s_{i, t}, a_{i, t}, \gain_{i, t})$.
Drawing from the online learning literature, each selector maintains an adaptive parameter vector $\theta_{i, t}$ that updates after each round based on observed context $\ctx_{i, t}$ and game outcome $\xi_{i, t}$:
\begin{equation}
    s_{i, t} \sim \sigma_i(\cdot \mid \ctx_{i, t} ; \theta_{i,t}), \qquad
    \theta_{i,t+1} = \textsc{Update}(\theta_{i, t}, \ctx_{i, t}, \xi_{i, t}).
\end{equation}

Any architecture that maps $\theta_{i, t}$ to a selection policy $\sigma_{i,t}$ and supports updates from online feedback qualifies as a backbone.
Such options can include classical multi-armed bandit samplers \citep{ThompsonSampler}, neural network-based classifiers \citep{RouterDC, CausalLLMRouting}, or at the frontier, an LLM prompted with game history.
Consequently, $\theta_{i, t}$'s definition depends on the backbone, such as the network weights for a classifier, or the per-arm posterior for a Thompson sampler.
We note that the game outcome is also incorporated into the next round's context $\ctx_{i, t+1} = f(\ctx_{i, t}, \xi_{i, t})$ to be used in base game decision-making.
Finally, this is repeated over $N$ independent seed runs (where the parameter vector is reset for each seed $\theta_{i,1} \leftarrow \theta_i^{0}$) to account for stochastic variance in strategic play.
\Cref{alg:strategic_selector} summarises the overall procedure. Note that from player $i$'s view, the opponent's selector $\sigma_{-i}$, its fielded model $s_{-i, t}$, and action $a_{-i, t}$ are never observed. They are resolved by the game environment $\game$.

\begin{wrapfigure}{r}{0.55\textwidth}
    \begin{algorithm2e}[H]
    \caption{Repeated strategic selection, deployer $i$'s view, in game $\game$, across $N$ seed runs} \label{alg:strategic_selector}
    \begin{spacing}{0.7}
    \small
    Initialize $S_i=\{1,\dots,M_i\}$, $\theta_i^{0}$, $T$ rounds, $N$ seeds;
    \For{$n \leftarrow 1$ \KwTo $N$}{
        \tcp{runs are independent}
        $\theta_{i,1} \leftarrow \theta_i^{0}$, $\ctx_{i, 1} \in \Xcal$\;
        \For{$t \leftarrow 1$ \KwTo $T$}{
            $s_{i,t} \sim \sigma_i(\cdot \mid \ctx_{i, t}; \theta_{i,t})$\;
            \tcp{model takes action}
            $a_{i, t} \leftarrow s_{i, t}(x_{i, t})$\; %
            \tcp{base game resolves}
            $\val_{i, t}, \cost_{i, t}, g_{i, t} \leftarrow \game(a_{i, t})$\;
            $\xi_{i, t} \leftarrow (s_{i,t}, a_{i,t}, \gain_{i,t})$\;
            \tcp{selector parameter is updated}
            $\theta_{i,t+1} \leftarrow \textsc{Update}(\theta_{i,t}, \ctx_{i, t}, \xi_{i, t})$\;
            $\ctx_{i, t+1} \leftarrow f(\ctx_{i, t}, \xi_{i, t})$\;
        }
    }
    \end{spacing}
    \end{algorithm2e}
\end{wrapfigure}

\subsection{Noise in Strategic Play}
Single-trial LLM decision-making is noisy, subject to the probabilistic sampling mechanism, and sensitive to variations in prompts.
Related work addresses LLM evaluations by adapting cross-disciplinary methods such as signal processing \citep{Heinemann}, experiment-design theory \citep{Miller}, and psychometrics \citep{StocEvals}.
They find that in traditional evaluations (e.g. question-answering benchmarks), even a modest amount of replication (e.g. $N\approx5$) substantially stabilises the estimates.
We argue that due to its interactive opponent-conditioned structure, strategic LLM evaluation is inherently noisier and requires a higher number of replications.
To quantify the noise in strategic play, we define a method to measure the levels of signal and noise across games by adapting \citet{Heinemann}'s work.
Building on this, we develop approaches to choose $N$ for a given game-portfolio pair \citep{Miller}.
See Appendix~\ref{app:evaluation-signal} for further details.

\subsection{Evaluation Metrics} 
To measure strategic performance, we capture the game utility $\val_i$, associated cost $\cost_i$, and their trade-off represented as the net gain $\gain_i$.
The game utility $\val_i$ is min-max normalised ($\hat{\val}_i \in [0, 1]$) based on its game-specific bounds (e.g. the economic profit in Cournot Duopoly, or the number of winning battlefields in Colonel Blotto game).
The associated inference cost $\cost_i$ is represented by the per-model token usage cost (in dollars \$) using the OpenRouter\footnote{\url{https://openrouter.ai/models}} listings as market proxies.
This cost is min-max normalised ($\hat{\cost}_i \in [0, 1]$) based on the portfolio-bounds (minimum and maximum per-token price observed across the model portfolio).
Finally, we define net gain $\gain_i$ as the weighted sum of normalised utility $\hat{\val}_i$ and cost $\hat{\cost}_i$
\begin{equation}
\gain_i = \frac{\hat{\val}_i + \lambda (1 - \hat{\cost}_i)}{1 + \lambda}, \qquad \gain_i \in [0, 1],
\end{equation}
where $\lambda$ represents the cost trade-off weight and is fixed at $\lambda=1$ across our runs.
Any game episode is replicated across $N$ seeds to account for stochastic noise.
Each of the $N$ seed runs yields a single sample, from which we compute the mean $\bar{X}$, the standard error (SE), and the 95\% confidence interval (CI).
Alternative metrics include the in-game winning rate (i.e. we ignore the raw $\gain_i$ and measure $\mathbbm{1}[\gain_i > \gain_{-i}]$), or representing the cost $\cost_i$ by the per-model throughput (tokens produced per second).

\section{Experiments}
\label{sec:experiments}
We provide empirical evidence in 3 steps and motivate the problem of strategic LLM selection.
First, an exhaustive pairwise matchup across 10 models and 5 games motivates the problem, showing that strategic capability is varied and exploitable by a selector.
Following these findings, a suite of selectors is run against fixed but unknown models (i.e. opponent is a fixed LLM) to evaluate their potential as strategic selectors.
Finally, these selectors are deployed against adaptive opponents to simulate the non-stationary dynamics of real-world settings.
We note that reported results should not be compared across (\Cref{tab:fixed_model_results,tab:selectors_vs_fixed,tab:selector_vs_selector_cournot}), as a method's performance is a function of the specific opponent pool it faced which varies across settings (e.g. the runs against adaptive opponents do not include fixed models).

\subsection{Design}

Full details on the models, games, and selector implementations are given in Appendix~\ref{appendix:setup}.

\textbf{Models and Games.}
The model portfolio draws on 10 open-weight LLMs across four families (Granite, Mistral, Llama, and Qwen) ranging from 3B to 70B to capture varying capability levels, usage costs, and dispositions.
The game inventory captures 5 classical game-theoretic scenarios (Colonel Blotto, Cournot Duopoly, Stag Hunt, Vickrey Auction, Chicken Game) spanning zero-sum to cooperative dilemmas.
Experiments are run through OutplayArena\footnote{\url{https://github.com/OutplayArena/arena}}, which owns the base game rules, turn sequencing, and scoring.

\textbf{Selectors.}
A suite of selectors spanning traditional and adversarial bandit samplers \citep{ThompsonSampler, EXP3Paper}, contemporary LLM routers \citep{RouterDC, CausalLLMRouting}, and an LLM self-reflection framework \citep{ReflexionFramework} is then incorporated to draw on this portfolio.
Adapting selectors from a range of different backgrounds plays an important diagnostic role in understanding the strengths and weaknesses of each method as strategic selectors.
Additionally, simple opponent-aware full-information selectors are evaluated post-hoc to measure the selection headroom.

\textbf{Number of Rounds and Seeds.}
Some strategic benchmarks \citep{TMGBench} evaluate models across hundreds of games.
To test whether such broad evaluations rest on a consistent signal, we quantify the per-game signal-to-noise ratio (SNR) and find that the evaluation signal is inherently game-specific.
A power analysis on single-round fixed-model matchups shows that the share of detectable pairwise comparisons stabilises at $N \approx 150$ seeds in every evaluated game, with Cournot Duopoly separating models more cleanly than any other game (Appendix~\ref{app:evaluation-signal-analysis}).
In our experiments, fixed-model matchups are single-round $T=1$ with $N=300$ seeds.
Selectors learn online and require repeated play, so we run $T=300$ rounds per seed with a single seed ($N=1$) to keep the inference cost tractable.
Note that the fielded model's and fixed opponent's context is limited to the current round.
This ensures that each round is an independent draw of the opponent’s play.

\subsection{Results}

\begin{figure*}[t]
\hfill
\begin{subfigure}[b]{0.45\textwidth}
\vspace{0pt}
\centering
\includegraphics[width=1\linewidth]{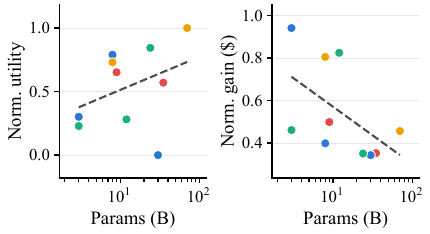}\vspace{-0.5em}
\caption{Normalised $\val$~$\uparrow$ and $\gain$(\$)~$\uparrow$ across all games.}
\label{fig:fragmented_scale}
\end{subfigure}
\hfill
\begin{subfigure}[b]{0.45\textwidth}
\vspace{0pt}
\centering
\includegraphics[width=1\linewidth]{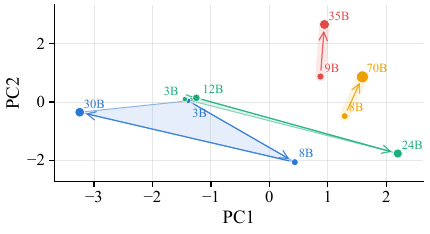}\vspace{-0.5em}
\caption{Model dispositions on the first two PCs.}
\label{fig:fragmented_disposition}
\end{subfigure}
\caption{\emph{(a)} Model scale does not reliably translate into strategic capability.
\emph{(b)} Models differ in strategic disposition.
Each point shows a model at the projection of its seven in-game disposition scores onto the first two principal components. \textcolor{granitecolor}{\textbf{Granite}}, \textcolor{mistralcolor}{\textbf{Mistral}}, \textcolor{llamacolor}{\textbf{Llama}}, \textcolor{qwencolor}{\textbf{Qwen3.5}}.
}
\label{fig:selection_needed}
\vspace{-0.1em}
\end{figure*}

\textbf{Strategic Capabilities Vary Across Models and Games.}
\Cref{tab:fixed_model_results} reveals that strategic competence is uneven and game dependent.
For example, the Llama models are top performers in Cournot Duopoly but fail to make the podium in Chicken Game.
In addition, the substantial performance variance among models within individual games highlights an opportunity to adopt a selector.
\Cref{fig:fragmented_scale} underscores that model scale does not reliably translate into strategic capability. 
Across our evaluations, we find only a weak positive correlation between model scale and utility (Pearson's $r=0.37$), and a moderate negative relationship when accounting for the usage costs ($r=-0.54$).
Upon a closer look, we also see that smaller models have game-specific usability.
For example in Vickrey Auction, due to its high rationality (i.e. always bidding the true value), the Granite4.1 3B outperforms other much bigger models that range from 24B to 70B parameters.
Finally, \cref{fig:fragmented_disposition} supports our framing ``choosing a model is choosing a strategy''.
For each model, we score strategic dispositions from its in-game actions, such as the dare rate in Chicken Game or the produced quantity in Cournot Duopoly.
We describe each model by a vector of its seven disposition scores, and project these vectors onto their principal components (Appendix~\ref{app:fixed_disposition}).
We find that even same family models can differ sharply in disposition.
For example, Granite4.1 8B dares in 96\% of Chicken Game rounds while its 3B and 30B versions never dare.
Families still show a general coherence, as Qwen3.5 and Llama models compete aggressively in Cournot Duopoly, whereas most Granite and Mistral models cooperate.
We believe this is an important signal that a strategic selector should learn to infer and exploit.

\begin{table}[t]
\centering
\footnotesize
\setlength{\tabcolsep}{5pt}
\renewcommand{\arraystretch}{1.05}
\caption{Normalised in-game utility $\val\uparrow$ acquired by each model across 5 games. 
We report the mean $\pm$ 95\% CI, $\times100$, aggregated across all pairwise matchups, using $T=1$ round, $N=300$ seeds per run, 
and \colorbox{aliceblue}{\textbf{highlight}} the best score for each game.}
\resizebox{0.9\linewidth}{!}{%
\begin{tabular}{lrrrrr}
\toprule
\rowcolor{lavenderblush}\textbf{Model} & \textbf{Chicken Game} & \textbf{Colonel Blotto} & \textbf{Cournot Duopoly} & \textbf{Stag Hunt} & \textbf{Vickrey Auction} \\
\midrule
Granite4.1 3B          & \cellcolor{aliceblue}\textbf{87.5 $\pm$ 0.3} & 50.6 $\pm$ 0.3 & 26.0 $\pm$ 1.2 & 74.3 $\pm$ 2.9 & \cellcolor{aliceblue}\textbf{74.3 $\pm$ 0.6} \\
Granite4.1 8B          & 69.6 $\pm$ 3.0 & 50.5 $\pm$ 0.3 & 7.8 $\pm$ 1.5  & 65.6 $\pm$ 2.2 & 73.7 $\pm$ 0.6 \\
Granite4.1 30B         & 87.3 $\pm$ 0.3 & 50.4 $\pm$ 0.2 & 13.2 $\pm$ 1.5 & 72.9 $\pm$ 2.9 & 69.4 $\pm$ 0.5 \\
Ministral3 3B        & 86.4 $\pm$ 0.7 & 48.5 $\pm$ 0.5 & 18.4 $\pm$ 2.0 & 73.0 $\pm$ 2.8 & 72.0 $\pm$ 0.6 \\
Mistral-Nemo 12B    & 84.2 $\pm$ 1.2 & 50.3 $\pm$ 0.2 & 20.0 $\pm$ 2.3 & 72.7 $\pm$ 2.8 & 71.9 $\pm$ 0.6 \\
Mistral-Small 3.2 24B   & 82.2 $\pm$ 1.7 & 49.7 $\pm$ 0.3 & 35.2 $\pm$ 2.8 & 56.1 $\pm$ 1.2 & 72.5 $\pm$ 0.6 \\
Llama3.1 8B        & 77.1 $\pm$ 2.3 & 49.8 $\pm$ 0.4 & 40.4 $\pm$ 2.7 & 64.5 $\pm$ 2.0 & 71.5 $\pm$ 0.7 \\
Llama3.3 70B       & 69.9 $\pm$ 2.9 & 50.4 $\pm$ 0.2 & \cellcolor{aliceblue}\textbf{48.9 $\pm$ 2.9} & 75.7 $\pm$ 2.6 & 73.8 $\pm$ 0.6 \\
Qwen3.5 9B  & 72.0 $\pm$ 2.8 & 49.6 $\pm$ 0.4 & 23.2 $\pm$ 2.3 & 74.7 $\pm$ 2.6 & 73.0 $\pm$ 0.6 \\
Qwen3.5 35B & 83.7 $\pm$ 1.4 & 50.3 $\pm$ 0.3 & 42.0 $\pm$ 2.8 & \cellcolor{aliceblue}\textbf{77.2 $\pm$ 2.6} & 71.4 $\pm$ 0.7 \\
\bottomrule
\end{tabular}}
\label{tab:fixed_model_results}
\end{table}

\textbf{Strategic Context Needs Special Treatment.}
We first simplify our selection problem by playing against a fixed unknown opponent, to try and infer strategic learning signals against a stationary fixed LLM.
\Cref{tab:selectors_vs_fixed} indicates that not all traditional methods transfer equally to the strategic selection setup, with the LLM Selector scoring the lowest overall.
Upon closer inspection, we find that it also falls behind nearly every fixed model across most games (Appendix~\ref{app:selector_vs_fixed}).
The Causal LLM Router and RouterDC show relative improvement but are still beaten by several fixed models in multiple games.
We attribute this to the fact that strategic settings require special treatment of the game context and the learning signal.
Methods that rely on query embeddings for semantic similarity matching (Causal LLM Router, RouterDC), or on LLM reflection (LLM Selector), fail to surface implicit strategic signals.
A simple bandit algorithm using Thompson sampling achieves the highest net gain among all selectors in every game, despite relying on nothing more than per-round outcome feedback.
However, even in Cournot Duopoly, where the Thompson sampler proves most dominant, it fails to beat the best fixed model (Llama3.1 8B), indicating significant room for improvement.
We believe a better selector algorithm should learn to directly infer the latent, hidden per-game strategic signal, rather than relying solely on observed game outcomes.

\begin{table}[t]
\centering
\footnotesize
\setlength{\tabcolsep}{5pt}
\renewcommand{\arraystretch}{1.05}
\caption{Normalised net gain $\gain\uparrow$ (top) and in-game utility $\val\uparrow$ (bottom) acquired by each selector, against fixed opponents and across 5 games.
We show the mean $\pm$ 95\% CI, $\times100$, averaged across 10 runs (the size of our portfolio), using $T=300$ rounds, $N=1$ seed per run, 
and \colorbox{aliceblue}{\textbf{highlight}} the best score for each game.
Selectors' learning signal was set as their per-round gain $g_{i, t}$.
}
\resizebox{\linewidth}{!}{%
\begin{tabular}{lrrrrr}
\toprule
\rowcolor{oldlace}\textbf{Selector} & \textbf{Chicken Game} & \textbf{Colonel Blotto} & \textbf{Cournot Duopoly} & \textbf{Stag Hunt} & \textbf{Vickrey Auction} \\
\midrule
\multicolumn{6}{c}{\textit{Reports normalised gain $\gain$~$\uparrow$}} \\
\midrule
\textit{Per-Opponent Full-Information} & 92.8 $\pm$ 0.4 & 73.0 $\pm$ 0.2 & 80.7 $\pm$ 1.5 & 92.2 $\pm$ 1.0 & 86.7 $\pm$ 0.5 \\
\midrule
Thompson Sampler (TS)   & \cellcolor{aliceblue}\textbf{85.1 $\pm$ 0.5} & \cellcolor{aliceblue}\textbf{71.2 $\pm$ 0.2} & \cellcolor{aliceblue}\textbf{47.7 $\pm$ 0.8} & \cellcolor{aliceblue}\textbf{76.1 $\pm$ 0.8} & \cellcolor{aliceblue}\textbf{74.8 $\pm$ 0.1} \\
RouterDC                & 84.3 $\pm$ 0.1 & 68.6 $\pm$ 0.1 & 36.2 $\pm$ 0.1 & 74.4 $\pm$ 0.8 & 72.8 $\pm$ 0.2 \\
Causal LLM Router      & 78.6 $\pm$ 0.7 & 63.2 $\pm$ 0.5 & 31.3 $\pm$ 0.7 & 74.0 $\pm$ 0.9 & 64.8 $\pm$ 0.5 \\
LLM Selector            & 61.0 $\pm$ 0.6 & 46.9 $\pm$ 0.3 & 26.9 $\pm$ 0.5 & 46.6 $\pm$ 1.2 & 54.4 $\pm$ 0.2 \\
\midrule
\multicolumn{6}{c}{\textit{Reports normalised utility $\val$~$\uparrow$}} \\
\midrule
\textit{Per-Opponent Full-Information} & 88.9 $\pm$ 0.7 & 50.6 $\pm$ 0.3 & 43.0 $\pm$ 2.6 & 88.1 $\pm$ 2.1 & 75.1 $\pm$ 0.7 \\
\midrule
Thompson Sampler (TS)  & 87.4 $\pm$ 0.7 & 50.4 $\pm$ 0.1 & \cellcolor{aliceblue}\textbf{39.4 $\pm$ 1.4} & 78.0 $\pm$ 1.3 & \cellcolor{aliceblue}\textbf{73.4 $\pm$ 0.3} \\
RouterDC                & \cellcolor{aliceblue}\textbf{87.8 $\pm$ 0.2} & 50.5 $\pm$ 0.1 & 19.6 $\pm$ 0.3 & 76.1 $\pm$ 1.5 & 72.8 $\pm$ 0.3 \\
Causal LLM Router      & 79.7 $\pm$ 1.1 & 50.1 $\pm$ 0.2 & 27.1 $\pm$ 1.3 & 72.4 $\pm$ 1.4 & 72.2 $\pm$ 0.4 \\
LLM Selector            & 80.6 $\pm$ 1.2 & 50.1 $\pm$ 0.2 & 18.1 $\pm$ 1.0 & \cellcolor{aliceblue}\textbf{87.0 $\pm$ 2.3} & 71.5 $\pm$ 0.5 \\
\bottomrule
\end{tabular}}
\label{tab:selectors_vs_fixed}
\end{table}

\textbf{An Interactive Learning Problem.}
In practice, opponents can also use a selector that adapts its policy each round based on the observed game outcome.
To capture this setting, we extend our evaluations and play against adaptive selector opponents.
\Cref{tab:selector_vs_selector_cournot} reveals that multi-armed bandit algorithms remain dominant.
However, the non-stationary adaptations (Sliding-Window TS and Discounted TS) fail to exploit the underlying concept drift in adaptive settings to outclass the plain TS.
Similarly, an adversarial variant (EXP3) performs noticeably worse, on par with embedding-based methods.
Hence, we note that simple multi-armed bandit adaptations are not sufficient to capture the implicit learning signals shaped by repeated strategic interaction.

\begin{table}[t]
\centering
\footnotesize
\caption{Normalised net gain $\gain\uparrow$, in-game utility $\val\uparrow$, and raw usage cost $\cost\downarrow$ acquired by each selector, against other selectors and in Cournot Duopoly.
We report the mean $\pm$ 95\% CI, $\times100$, aggregated across all pairwise matchups, using $T=300$ rounds, $N=1$ seed per run, 
and \colorbox{aliceblue}{\textbf{highlight}} the best score for each metric.
Selectors' learning signal was set as their per-round gain $g_{i, t}$.}
\label{tab:selector_vs_selector_cournot}
\setlength{\tabcolsep}{5pt}
\renewcommand{\arraystretch}{1.05}
\resizebox{0.8\linewidth}{!}{%
\begin{tabular}{lrrr}
\toprule
\rowcolor{mintcream}\textbf{Selector} & \textbf{net gain $\gain$}~$\uparrow$ & \textbf{in-game utility $\val$}~$\uparrow$ & \textbf{usage cost $\cost$ ($\times10^{-3}$\$)}~$\downarrow$ \\
\midrule
Thompson Sampler (TS)       & \cellcolor{aliceblue}\textbf{72.4 $\pm$ 0.8} & \cellcolor{aliceblue}\textbf{47.1 $\pm$ 1.7} & 0.027 $\pm$ 0.001 \\
Sliding-Window TS     & 65.9 $\pm$ 0.7 & 33.2 $\pm$ 1.4 & 0.017 $\pm$ 0.001 \\
Discounted TS    & 63.1 $\pm$ 0.8 & 27.5 $\pm$ 1.6 & 0.017 $\pm$ 0.001 \\
EXP3                  & 60.2 $\pm$ 0.7 & 22.1 $\pm$ 1.4 & 0.020 $\pm$ 0.001 \\
RouterDC              & 59.6 $\pm$ 0.3 & 19.3 $\pm$ 0.5 & \cellcolor{aliceblue}\textbf{0.002 $\pm$ 0.000} \\
Causal LLM Router          & 59.1 $\pm$ 0.7 & 19.7 $\pm$ 1.3 & 0.019 $\pm$ 0.001 \\
LLM Selector           & 40.9 $\pm$ 0.7 & 19.6 $\pm$ 1.3 & 0.409 $\pm$ 0.004 \\
\bottomrule
\end{tabular}}
\end{table}

\section{Lessons Learned}
\label{sec:lessons-learned}

\subsection{How to Learn from Strategic Context?} 
We believe strategic context is multi-faceted and encapsulates the game instructions (i.e. the strategic scenario we are dealing with), the portfolio model dispositions (i.e. understanding our resources to act on this scenario), and the opponent's actions that shape our scenario.
Together, this context is either available as a prior (e.g. through existing game-specific runs across models), or gathered post-hoc by observing the game outcome $\xi_{i, t}$ at the end of each round.
Traditional routing and sampling methods fail to make best use of this context.
Existing methods either encode the entire context (e.g. game instructions, history) as a whole, as form of vector embedding (RouterDC), without an underlying form or structure, or rely entirely on the observed game outcome (Thompson Sampler) with no regard to the strategic scenario (captured by the game instructions) we are dealing with.
A better selector should infer game-specific strategic cues, and map selection choices to their implications for future game states.
It should build an understanding of the performance of its own portfolio models, and the opponent under the current scenario.

\subsection{Games as Strategic Testbeds}
We saw that best-performing models vary noticeably across the 5 evaluated games, each setup capturing a particular strategic scenario (a model-to-model interplay) (\Cref{tab:fixed_model_results}).
Moreover, an observation can be made that games vary in difficulty (i.e. complexity).
For example, one can say that identifying the equilibrium strategy would require higher model reasoning capabilities in a game like Cournot Duopoly (an economic profit game) than in Colonel Blotto, which relies on a simple per-battlefield resolution rule (the larger number wins).
Empirically, this idea is reflected in the correlation between model scale and performance, which is the highest in Cournot Duopoly (Pearson's $r=0.66$), and much lower in games like Stag Hunt ($r=0.12$) and Colonel Blotto ($r=0.24$) (\Cref{fig:detailed-scale}).
Mathematically, there are an infinite number of normal- and extensive-form games that range in complexity and the strategic interplay captured.
Future work should therefore establish well-defined methods to classify a game's difficulty and strategic value.
Such methods are of high value to deployers before they decide where to train and use their agent workflows, as well as towards designing strategic selectors that better understand the underlying strategic learning signal.

\subsection{Games as Dynamic Testbeds}
Real-world AI deployments are messy. 
Inputs are noisy, and the agents involved often hold different, sometimes conflicting objectives.
Our results show that a model's competence is shaped by the opponent it faces, and how both sides adapt as the interaction unfolds.
Traditional benchmarks fail to capture this, as they test one model at a time against a fixed correct answer.
This makes game inventories a natural fit for evaluating models in decision-pressured, interactive settings that they are actually deployed to in practice.
Moreover, because game states are generated dynamically through interaction, games are more resistant to the training-data contamination that increasingly undermines static benchmarks \citep{BDCMitigate}.

That said, games are not immune to saturation.
Even without contamination, the discriminative signal can erode as models approach equilibrium play (strategy profiles from which no player can gain by unilaterally deviating).
More specifically, under standard no-regret learning, adaptive play converges to the set of coarse correlated equilibria (CCE), a superset of Nash equilibria \citep{AlgorithmicGameTheory}.
In two-player zero-sum games such as Colonel Blotto, every equilibrium yields the same expected payoff, so once models play near-equilibrium strategies, none can outperform another and their results become indistinguishable.
In contrast, in general-sum games the set of CCE is typically large and contains outcomes with very different payoffs.
This suggests favouring general-sum games when building inventories meant to remain discriminative.

\section{Conclusion and Future Work}
\label{conclusion}

We extend game-theoretic evaluations and define strategic selection as a repeated opponent-conditioned meta game layered on top of the base game.
Initial evaluations across 5 games and 10 models show that LLM strategic capability varies substantially across games and models, leaving considerable room for a selector to exploit.
By adapting a range of selection methods, we report that the existing methods do not transfer equally well even against a simple fixed-model opponent, and that a plain bandit algorithm can surpass structurally heavier pre-trained encoders and LLM reflection methods.
Subsequent runs against a selector opponent reveal that simple adaptations like discounted and adversarial bandit algorithms are not able to capture the non-stationary dynamics of opponent-conditioned strategic selection, and fail to outclass the plain sampler.
Overall, the results highlight the challenges of strategic model selection, and motivate a selector design that can extract and learn from signals across 3 different dimensions: model capability, usage cost, and strategic disposition, in noise-heavy interactive settings.

Our investigation was scoped to the analysis of 5 games.
However, to better understand agent-to-agent interplay and the underlying safety and risk factors, it is important to consider games from a range of fields, including economics, social sciences, and operations research.
We believe this choice should reflect both the measurable effect of the game and the underlying practical application of the selector.
We also relied on explicit parsed game actions as a communication method between the models, which underestimates both the strategic capabilities (e.g. persuasion), and the associated risk factors in deployment.
A more realistic setting should capture agent-to-agent natural language interactions \citep{GameTalk}.
We leave such free-form interactions to future work.
Finally, high-stakes repeated interactions (e.g. negotiations, auctions) involve a risk factor not captured by expected gain.
This can include catastrophic losses, regulatory penalties, or reputational damage.
Capturing the variance in realised gain to account for involved risk is another future direction.
For example, a risk knob ($\beta\in\mathbb{R}$: $\beta=0$ is risk-neutral, $\beta>0$ risk-averse, $\beta<0$ risk-seeking) could allow users to better align the selector with their deployment policies.

\bibliographystyle{abbrvnat}
\bibliography{references}

\clearpage
\input{appendix.tex}

\end{document}

%% file: appendix.tex
\newpage
\appendix
\counterwithin{figure}{section}
\counterwithin{table}{section}

\begin{center}
    {\bf\Large Appendix}
\end{center}

\startcontents[sections]
\printcontents[sections]{l}{1}{\setcounter{tocdepth}{2}}
\newpage

\section{Glossary of Notation}
\label{sec:glossary}

\begin{table}[H]
\centering
\footnotesize
\caption{Glossary of notation.}
\begin{tabular}{@{}ll@{}}
\toprule
\textbf{Symbol} & \textbf{Meaning} \\
\midrule
\multicolumn{2}{@{}l}{\textbf{Meta Game} {$\Gamma$} (LLM selection game)} \\
\midrule
$i, -i$                                      & player indices, where $-i$ denotes the opponent of $i$ \\
$t = 1, \dots, T$                            & round index of the game \\
$N$                                          & number of seeds in the game (independent replications) \\
$S_i = \{1, \dots, M_i\}$                    & finite set of pure strategies, i.e. player $i$'s model portfolio \\
$\sigma_i \in \simplex(S_i)$                 & mixed strategy for player $i$, a probability simplex over $S_i$ \\
$\ctx_{i, t} \in \Xcal$                           & context observed by player $i$ before round $t$ \\
$\xi_{i, t}$                                      & game outcome produced by player $i$ on round $t$ \\
$s_{i,t} \sim \sigma_i$                      & pure strategy (model choice) fielded by player $i$ at round $t$ \\
$\val_{i,t}$                                 & utility obtained by player $i$ from the base-game outcome \\
$\cost_{i,t}$                                & inference cost incurred by player $i$ (tok/\$) \\
$\gain_{i,t}$     & net gain that captures the utility-cost trade-off \\
$\theta_{i,t}$                               & internal state of player $i$'s selector at round $t$ \\
\midrule
\multicolumn{2}{@{}l}{\textbf{Base Game} $\mathcal{G}$ (e.g. Colonel Blotto, Cournot Duopoly)} \\
\midrule
$a_{i,t}$                                    & action produced by the fielded model $s_{i,t}$ at round $t$ \\
\bottomrule
\end{tabular}
\label{tab:notation}
\end{table}

\section{More Details about Noise in Strategic Play}
\label{app:evaluation-signal}

\subsection{Signal and Noise Ratio}
\label{app:snr}
Following \citet{Heinemann}, we first define a descriptive method that characterises signal and noise levels across games.
Fix a game $\game$ and let $S$ denote its set of models under evaluation, with $j,k \in S$ ranging over the pair of models compared in $\game$.
For a model $s \in S$, let $\tau = 1, \dots, N$ index its seeds, where a seed is a single independent evaluation run.
Let $\val_{\game,s,\tau}$ be the bounds-normalised score on seed $\tau$ of $\game$ played by model $s$.
Let $\overline{\val}_{\game,s}$ denote the mean of $\val_{\game,s,\tau}$ over all $N$ seeds, and let $\overline{\val}_{\game}$ denote the mean of $\overline{\val}_{\game,s}$ over all models $s \in S$.
Noise, signal, and signal-to-noise ratio (SNR) are then
\begin{equation}
\text{noise}(\game,s) = \frac{\text{Std}_\tau(\val_{\game,s,\tau})}{\overline{\val}_{\game,s}}, \qquad
\text{signal}(\game) = \frac{\max_{j,k \in S}|\overline{\val}_{\game,j}-\overline{\val}_{\game,k}|}{\overline{\val}_{\game}}, \qquad
\text{SNR}(\game) = \frac{\text{signal}(\game)}{\overline{\text{noise}}(\game)}.
\label{eq:snr}
\end{equation}
Here $\text{Std}_\tau(\cdot)$ denotes the standard deviation taken over the seed index $\tau$, and $\overline{\text{noise}}(\game)$ denotes $\text{noise}(\game,s)$ averaged over every model $s \in S$.
Noise is a coefficient of variation, a model's own seed-to-seed spread relative to its mean.
Signal is the largest gap between any two models' means, relative to the game's overall mean.
SNR is high when models are well separated relative to how much a single model's score jitters across seeds, and low when the two quantities are comparable.

\subsection{Number of Seeds}
\label{appsub:number-of-seeds}
Given the SNR, we want to determine the minimum number of seeds $N$ that lets us report statistically significant outcomes.
This is a trade-off between cost (the number of model inferences we run) and reliability (our statistical confidence in the reported results).

\textbf{Decision accuracy.}
\citet{Heinemann} suggests calculating a decision accuracy metric to determine the number of seeds $N$ needed to resolve the pairwise gaps between models.
Let $\mathcal{P}$ denote the set of model pairs compared in $\game$, let $n$ denote a candidate (small) batch size, and let $N \gg n$ denote a large reference batch size.
Let $\overline{\val}_{\game,s}^{(n)}$ denote model $s$'s mean bounds-normalised score over $n$ seeds. Decision accuracy at batch size $n$ is then
\begin{equation}
    \mathrm{DA}(n) = \frac{1}{|\mathcal{P}|}\sum_{\{s_A,s_B\} \in \mathcal{P}} \mathbbm{1}\!\left[\, \mathrm{sign}\!\left(\overline{\val}_{\game,s_A}^{(n)}-\overline{\val}_{\game,s_B}^{(n)}\right) = \mathrm{sign}\!\left(\overline{\val}_{\game,s_A}^{(N)}-\overline{\val}_{\game,s_B}^{(N)}\right) \right],
\label{eq:da}
\end{equation}
the fraction of the pairs in $\mathcal{P}$ whose sign at batch size $n$ agrees with the sign at the reference batch size $N$.
Every $n$ below $N$ is estimated as the mean over repeated independent subsamples of the $N$-seed reference, drawn per model pair so each draw mirrors an actual $N$-seed experiment run at batch size $n$.
However, decision accuracy needs a reference batch, and in game-theoretic evaluations this reference is often not remotely the scale of traditional benchmarks, since the game-theoretic reference batch is self-generated instead of drawn from an external test suite.

\textbf{Minimum detectable effect.}
A more practical approach is to use the minimum detectable effect (MDE), which does not need a reference batch, and asks directly whether the current $N$ and opponent-pool size is sufficient to resolve the pairwise gaps we report.
Power analysis of \citet{Miller} (Eqs. 9-10), for two models $s_A$ and $s_B$ with $M$ shared opponents and $K$ seeds played against each opponent, gives the MDE as
\begin{equation}
    \delta(K) = (z_{\alpha/2} + z_\beta) \sqrt{\frac{\omega^2 + \sigma_{s_A}^2/K + \sigma_{s_B}^2/K}{M}},
\label{eq:mde}
\end{equation}
\begin{equation}
    \omega^2 = \mathrm{Var}_o\!\left(\overline{\val}_{\game,s_A,o} - \overline{\val}_{\game,s_B,o}\right), \qquad
    \sigma_{s_A}^2 = \mathbb{E}_o\!\left[\mathrm{Var}_t(\val_{\game,s_A,o,t})\right], \qquad
    \sigma_{s_B}^2 = \mathbb{E}_o\!\left[\mathrm{Var}_t(\val_{\game,s_B,o,t})\right].
\label{eq:mde-terms}
\end{equation}
Here $o$ ranges over the $M$ shared opponents ($M=8$ in a 10-model pool experiment), and $t=1,\dots,K$ indexes seeds played against a given opponent.
$z_\gamma$ denotes the $(1-\gamma)$ quantile of the standard normal distribution, so $\alpha=0.05$ (giving $z_{\alpha/2}$) and power $1-\beta=0.8$ (giving $z_\beta$) are \citet{Miller}'s own defaults for significance and power.
$\omega^2$ is the variance, across the $M$ shared opponents, of the cluster-mean gap between $s_A$ and $s_B$, i.e. how much that gap swings depending on the opponent, and $\sigma_{s_A}^2$ and $\sigma_{s_B}^2$ are each model's mean within-opponent variance across seeds.
All three quantities are estimated from a reference (e.g. $N=300$). $\delta(K)$ is the smallest true skill gap detectable at that $K$, for a fixed opponent-pool size.
For game $\game$ evaluated on a model pool $S$, the detectable fraction at $K$ seeds is
\begin{equation}
    \rho_\game(K) = \frac{1}{\binom{|S|}{2}}\sum_{\{s_A,s_B\} \subset S} \mathbbm{1}\!\left[\, |\overline{\val}_{\game,s_A}-\overline{\val}_{\game,s_B}| \geq \delta(K) \,\right],
\label{eq:mde-ratio}
\end{equation}
the share of the $\binom{|S|}{2}$ pairwise gaps in $\game$ that exceed $\delta(K)$. Sweeping $K$ per game produces a rough signal on the statistical significance of pairwise model comparisons from that game.

\subsection{Analysis}
\label{app:evaluation-signal-analysis}

\textbf{Scope of the analysis.}
All signal, noise, and seed-stabilisation analyses in this section are computed on fixed-model matchups, using single-round episodes ($T=1$). 
Each seed is therefore an independent draw of the base game, and the estimates characterise the inherent variability of strategic play between two given models. 
We note that our analyses do not directly transfer to selector runs, where selector's own choices can evolve over the rounds.

\begin{figure}[H]
    \centering
    \sbox{\tablebox}{%
        \footnotesize
        \setlength{\tabcolsep}{3pt}%
        \renewcommand{\arraystretch}{0.9}%
        \begin{tabular}{lrrr}
        \toprule
        Game & Signal & Noise & SNR \\
        \midrule
        \texttt{Cournot Duopoly} & \textbf{1.431} & \textbf{1.537} & \textbf{0.931} \\
        \texttt{Chicken Game}    & 0.224 & 0.357 & 0.627 \\
        \texttt{Stag Hunt}       & 0.315 & 0.559 & 0.564 \\
        \texttt{Vickrey Auction} & 0.066 & 0.129 & 0.512 \\
        \texttt{Colonel Blotto}  & 0.040 & 0.089 & 0.449 \\
        \bottomrule
        \end{tabular}}%
    \sbox{\chartbox}{\includegraphics[width=0.35\textwidth]{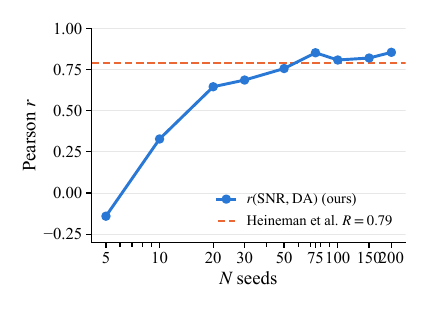}}%
    \begin{subfigure}[b]{\wd\tablebox}
        \parbox[c][\ht\chartbox][c]{\wd\tablebox}{\usebox{\tablebox}}
        \caption{}
        \label{tab:snr}
    \end{subfigure}%
    \hspace{2em}%
    \begin{subfigure}[b]{\wd\chartbox}
        \usebox{\chartbox}
        \caption{}
        \label{fig:snr_da_correlation}
    \end{subfigure}
    \caption{Statistical analysis of signal and noise across 5 games and 4 model families.
    \emph{(a)} Signal, noise, and SNR per game, $n=300$.
    \emph{(b)} Similar to \citet{Heinemann} $(R = 0.79)$ we show that our correlation holds at $r \approx 0.81$ from $n=75$ onwards.}
    \label{fig:results-signal-and-noise}
\end{figure}

Signal-to-noise ratio is not game-agnostic.
Cournot Duopoly separates models almost twice as cleanly as any other game, while Colonel Blotto carries the lowest signal (\Cref{tab:snr}).
A qualitative investigation shows that this stems from all models playing Colonel Blotto very similarly, all converging to an equilibrium of an even distribution across battlefields.
This SNR ranking validity is confirmed independently by decision accuracy (\Cref{fig:snr_da_correlation}). Holding the seed count $n$ fixed and correlating SNR against decision accuracy across the five games gives Pearson $r \approx 0.81$ to $0.85$ for $n \geq 75$, close to \citet{Heinemann}'s reported $R=0.79$, indicating that our SNR metric captures the same effect as theirs.

\begin{figure}[H]
    \centering
    \begin{subfigure}[b]{0.45\textwidth}
        \centering
        \includegraphics[width=1\linewidth]{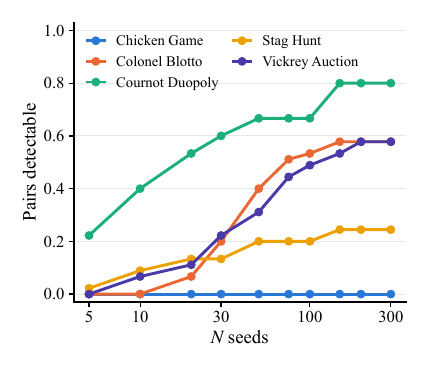}
        \caption{}
        \label{fig:mde}
    \end{subfigure}
    \begin{subfigure}[b]{0.45\textwidth}
        \centering
        \includegraphics[width=1\linewidth]{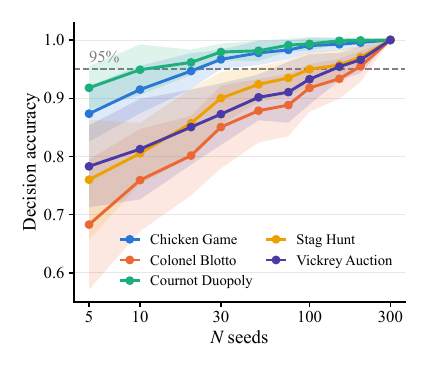}
        \caption{}
        \label{fig:da}
    \end{subfigure}
    \caption{Statistical analysis of signal and noise across 5 games and 4 model families. 
    \emph{(a)} Fraction of all 45 reported pairwise gaps that are statistically detectable.
    \emph{(b)}  Decision accuracy, mean over 30 subsample draws per $n$.}
    \label{fig:results-number-of-seeds}
\end{figure}

Decision accuracy results in \Cref{fig:da} follow the same ranking. Cournot Duopoly reaches 95\% decision accuracy at the smallest seed count of the five games, $n=20$, while Colonel Blotto needs the largest, $n=200$, consistent with its near-zero signal. No single seed count clears 95\% decision accuracy on every game before $n=200$.

In the power analysis (\Cref{fig:mde}), some games never clear the significance bar even at the full seed budget, while Cournot Duopoly has the highest ratio of detectable pairwise comparisons, confirming the ranking from both the SNR and the decision accuracy analysis.
The detection rate converges after around $\ge150$ seeds for the games that do clear the bar.
For the primary investigation, we fix the number of seeds to $N=300$.

\section{More Details about Experimental Setup}
\label{appendix:setup}

\subsection{Models}
\label{appendix:models}

Our experimental model portfolio spans 10 open-weight models with three size tiers, listed in \Cref{tab:portfolio}. Every model is served locally through Ollama.
Every model call uses a sampling temperature of 1.0 and nucleus sampling (top-p) of 0.9.
Every experiment runs on a workstation with 64 physical CPU cores (128 logical cores) and two NVIDIA RTX PRO 6000 Blackwell Workstation Edition GPUs, under Linux 7.0.0-28-generic (x86-64, glibc 2.39) with Python 3.12.3.

\begin{table}[H]
\centering
\footnotesize
\caption{Model portfolio used across experiments.}
\label{tab:portfolio}
\begin{tabular}{@{}ll@{}}
\toprule
\textbf{Model} & \textbf{Hugging Face repository} \\
\midrule
\texttt{granite4.1:3b} & \texttt{ibm-granite/granite-4.1-3b} \\
\texttt{granite4.1:8b} & \texttt{ibm-granite/granite-4.1-8b} \\
\texttt{granite4.1:30b} & \texttt{ibm-granite/granite-4.1-30b} \\
\texttt{llama3.1:8b-instruct-q4\_K\_M} & \texttt{meta-llama/Llama-3.1-8B-Instruct} \\
\texttt{llama3.3:70b} & \texttt{meta-llama/Llama-3.3-70B-Instruct} \\
\texttt{ministral-3:3b} & \texttt{mistralai/Ministral-3-3B-Instruct-2512} \\
\texttt{mistral-nemo:12b} & \texttt{mistralai/Mistral-Nemo-Instruct-2407} \\
\texttt{mistral-small3.2:24b} & \texttt{mistralai/Mistral-Small-3.2-24B-Instruct-2506} \\
\texttt{qwen3.5:9b} & \texttt{Qwen/Qwen3.5-9B} \\
\texttt{qwen3.5:35b} & \texttt{Qwen/Qwen3.5-35B-A3B} \\
\bottomrule
\end{tabular}
\end{table}

\subsection{Games}
\label{appendix:games}

The base game $\game$ is instantiated as one of five two-player games. 
For each game we have kept the default configurations provided by \href{https://github.com/OutplayArena/arena}{OutplayArena}.
We provide brief details about each game, and refer the reader to the source outlined in \Cref{tab:games} for further information.

\begin{itemize}
    \item \textbf{Colonel Blotto} is a zero-sum game where two players split a fixed budget across several battlefields. It measures how a model reallocates its strategy in response to an opponent's shifting distribution.
    \item \textbf{Cournot Duopoly} is a simultaneous quantity-setting game where two firms choose outputs that jointly fix market price and profit. It isolates a single continuous best-response signal that separates models by how closely each converges toward the equilibrium quantity.
    \item \textbf{Stag Hunt} is a coordination game with a risk-dominant and a payoff-dominant equilibrium. It measures whether a model favors the safer but lower-payoff option over the cooperative one when the opponent's action is uncertain.
    \item \textbf{Vickrey Auction} is a second-price sealed-bid auction where each bidder submits a value-based bid. It measures how closely a model's bid tracks its private valuation under the auction's truthful-bidding equilibrium.
    \item \textbf{Chicken Game} is an anti-coordination game built around brinkmanship, where mutual insistence is worst for both players. It measures a model's willingness to escalate or concede as the risk of mutual loss grows.
\end{itemize}

\begin{table}[H]
\centering
\footnotesize
\caption{The five base games $\game$ used across experiments, all played between 2 players.}
\label{tab:games}
\resizebox{\linewidth}{!}{%
\begin{tabular}{@{}lll@{}}
\toprule
\textbf{Game} & \textbf{Structure} & \textbf{Source} \\
\midrule
Chicken Game & anti-coordination, brinkmanship & \url{https://en.wikipedia.org/wiki/Chicken_(game)} \\
Colonel Blotto & zero-sum resource allocation & \url{https://en.wikipedia.org/wiki/Blotto_game} \\
Cournot Duopoly & simultaneous quantity competition & \url{https://en.wikipedia.org/wiki/Cournot_competition} \\
Stag Hunt & coordination, risk dominance & \url{https://en.wikipedia.org/wiki/Stag_hunt} \\
Vickrey Auction & second-price sealed-bid auction & \url{https://en.wikipedia.org/wiki/Vickrey_auction} \\
\bottomrule
\end{tabular}}
\end{table}

\subsection{Selectors}
\label{appendix:selectors}

We provide the implementation detail of each selector.

\textit{Fixed Model.} One fixed model across all seeds $N$ that keeps context and past outcomes across rounds $t = 1, \dots, T$ unless otherwise stated.

\textit{Per-Opponent Full-Information.} Given the identity of the fixed opponent, it looks up the round-robin table (our prior) once at construction and commits to whichever candidate scores highest against that opponent, fielding it every round with no further adaptation.

\textit{LLM Selector.} Uses \texttt{llama3.3:70b} as the selecting model and relies on the Reflexion framework \citep{ReflexionFramework} to pick the model to field. Namely, each round it is queried once with a description of the game, an entry per portfolio model (tier and per-token price), and a log of past rounds covering its own fielded model, action, and outcome signal. It parses a JSON response giving the picked model, a confidence, and a reasoning. Prior information enters only as in-context history, a rolling log of past rounds inside the prompt, with default sampling \texttt{temperature} = 1.0.

\textit{RouterDC.} Adapted from \citet{RouterDC}. Each round a frozen sentence encoder embeds the current query, built from player's own model, action, and signal history, and the pick is whichever candidate's trained embedding has highest cosine similarity to that query embedding. The embeddings come from a checkpoint fit offline, once per game. All prior information is folded into this pre-trained checkpoint at training time.

\textit{Causal LLM Router.} We follow the doubly robust regret-minimising rule of \citet{CausalLLMRouting}. Each round it draws a candidate from a softmax over its current per-(candidate, opponent) belief, with a floor that keeps every candidate at nonzero pick probability. It then folds in a doubly robust correction of the observed signal, using the pick probability it assigned as its own propensity instead of fitting a separate propensity model. We use the defaults \texttt{surrogate} = \texttt{softmax}, \texttt{tau} = 1.0, and \texttt{propensity\_floor} = 0.05.

\textit{Thompson Sampler.} It keeps a per-model Gaussian belief over the chosen signal that is refined as a running mean after every round. Each round it draws one sample per model from its current belief and fields the model with the highest draw. 
Additionally, we implement two non-stationary variants that change the belief update.
The discounted variant (\textit{Dis. Thompson Sampler}) multiplies every model's accumulated pseudo-count by the discount factor before folding in the new observation, so older evidence decays exponentially. The sliding-window variant (\textit{SW Thompson Sampler}) keeps only the $W$ (window size) most recent observations, and drops older ones from the belief entirely. We use \texttt{discount} = 0.99 in the discounted variant and \texttt{window} = 50 in the sliding-window variant.

\textit{EXP3.} Exponential-weights bandit updated by an importance-weighted multiplicative rule on the observed signal. Its mixed strategy blends the weight-proportional distribution with a uniform exploration floor, so every candidate keeps nonzero pick probability (\texttt{exploration\_rate} = 0.1).

\section{More Results}
\Cref{app:fixed_disposition} captures the complete in-game disposition analysis inferred from fixed model matchups.
\Cref{app:fixed_vs_fixed} gives the complete fixed-model-versus-fixed-model matchup results across all 5 games.
\Cref{app:selector_vs_fixed} then gives the full selector-versus-fixed-model matchup results.

\subsection{Fixed Model Disposition Scores}
\label{app:fixed_disposition}

For each game, we infer a small set of strategic dispositions from in-game play (\Cref{tab:dispositions}). 
To summarize how models differ, we standardize each disposition score across models and project the resulting vectors onto their two directions of largest variation. 
Together, these two principal components capture 69\% of the differences between models. 
We interpret PC1 as how competitive a model is and PC2 as how much it trusts its opponent.

\begin{table}[H]
\centering
\footnotesize
\setlength{\tabcolsep}{5pt}
\renewcommand{\arraystretch}{1.05}
\caption{Strategic disposition metrics and their loadings on the first two principal
components. All metrics are averaged across rounds and lie in $[0, 1]$.}
\label{tab:dispositions}
\resizebox{0.98\linewidth}{!}{%
\begin{tabular}{@{}llp{0.48\linewidth}rr@{}}
\toprule
\textbf{Game} & \textbf{Disposition} & \textbf{Definition} & \textbf{PC1} & \textbf{PC2} \\
\midrule
Chicken Game & Risk appetite & Fraction of rounds choosing dare over swerve.
  & $\mathbf{+0.35}$ & $-0.12$ \\
\midrule
\multirow{2}{*}{Cournot Duopoly}
  & Cooperativeness & Collusion index; 1 = joint-profit quantity, 0 = at or beyond Nash.
  & $\mathbf{-0.45}$ & $\mathbf{-0.41}$ \\
  & Aggressiveness  & Own minus opponent quantity, as a share of the maximum quantity.
  & $\mathbf{+0.49}$ & $\mathbf{+0.35}$ \\
\midrule
\multirow{2}{*}{Stag Hunt}
  & Risk appetite   & Fraction of rounds choosing stag over the safe hare.
  & $\mathbf{-0.38}$ & $\mathbf{+0.50}$ \\
  & Cooperativeness & Fraction of rounds landing on mutual stag.
  & $-0.34$          & $\mathbf{+0.54}$ \\
\midrule
\multirow{2}{*}{Vickrey Auction}
  & Risk appetite   & Mean own bid as a share of the maximum bid.
  & $\mathbf{+0.38}$ & $+0.29$ \\
  & Rationality     & One minus the mean $|\mathrm{bid} - \mathrm{value}|$; 1 = always bids its true value.
  & $+0.19$          & $-0.28$ \\
\midrule
\multicolumn{3}{@{}l}{Explained variance} & 43\% & 26\% \\
\bottomrule
\end{tabular}}
\end{table}

\begin{figure}[H]
  \centering
  \includegraphics[width=0.9\textwidth]{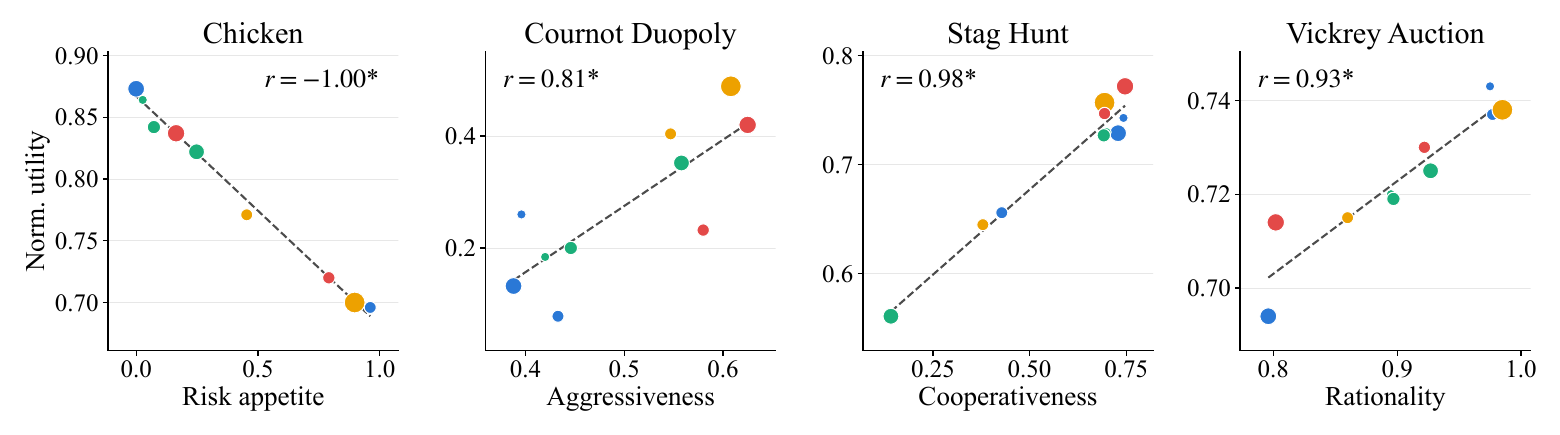} 
  \caption{Normalised in-game utility $\val\uparrow$ against the disposition. Dashed lines show the linear fit and $r$ its Pearson correlation. \textcolor{granitecolor}{\textbf{Granite}}, \textcolor{mistralcolor}{\textbf{Mistral}}, \textcolor{llamacolor}{\textbf{Llama}}, \textcolor{qwencolor}{\textbf{Qwen3.5}}.}
  \label{fig:disposition_vs_outcome}
\end{figure}

\begin{figure}[H]
  \centering
  \includegraphics[width=0.9\textwidth]{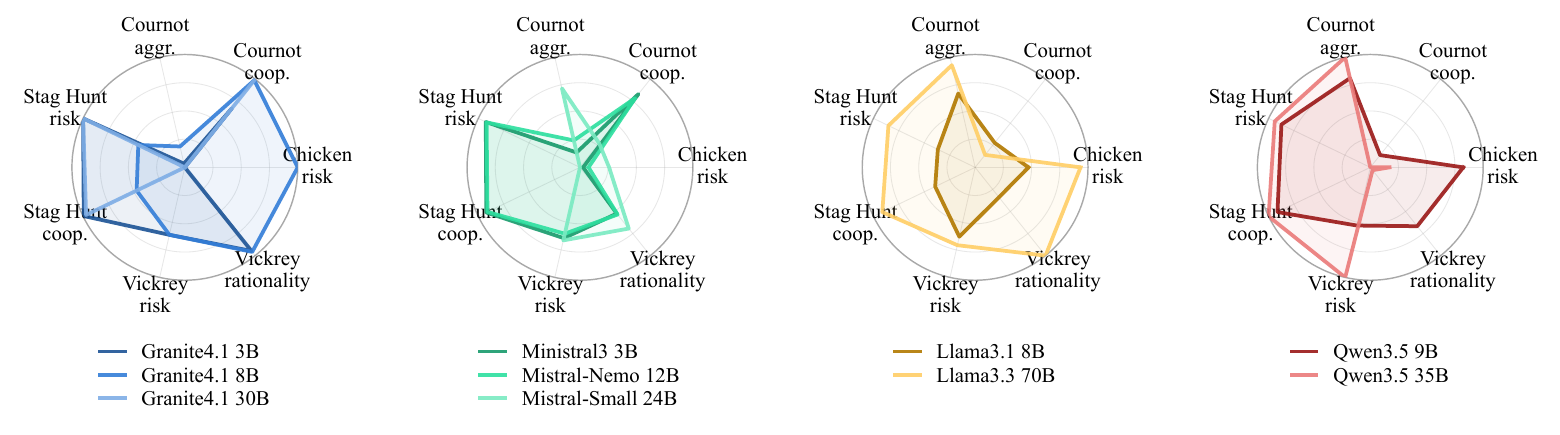} 
  \caption{Strategic dispositions inferred from in-game behaviour for each model family. Each axis is min-max normalised across the 10 models of the portfolio. \textcolor{granitecolor}{\textbf{Granite}}, \textcolor{mistralcolor}{\textbf{Mistral}}, \textcolor{llamacolor}{\textbf{Llama}}, \textcolor{qwencolor}{\textbf{Qwen3.5}}.}
  \label{fig:full_disposition}
\end{figure}

\subsection{Fixed Model vs. Fixed Model}
\label{app:fixed_vs_fixed}

\begin{table}[H]
\centering
\footnotesize
\setlength{\tabcolsep}{5pt}
\renewcommand{\arraystretch}{1.05}
\caption{Net gain, normalised utility, and normalised cost acquired by each model across 5 games. We report the mean $\pm$ 95\% CI, $\times100$, aggregated across all pairwise matchups, using $T=1$ round, $N=300$ seeds per run. \colorbox{aliceblue}{\textbf{Highlight}} marks the best score per game within each metric block.}
\resizebox{\linewidth}{!}{%
\begin{tabular}{lrrrrr}
\toprule
\rowcolor{lavenderblush}\textbf{Model} & \textbf{Chicken Game} & \textbf{Colonel Blotto} & \textbf{Cournot Duopoly} & \textbf{Stag Hunt} & \textbf{Vickrey Auction} \\
\midrule
\multicolumn{6}{c}{\textit{Reports normalised gain $\gain$~$\uparrow$}} \\
\midrule
Granite4.1 3B & \cellcolor{aliceblue}\textbf{93.8 $\pm$ 0.1} & \cellcolor{aliceblue}\textbf{75.2 $\pm$ 0.1} & 63.0 $\pm$ 0.6 & \cellcolor{aliceblue}\textbf{87.2 $\pm$ 1.4} & \cellcolor{aliceblue}\textbf{87.2 $\pm$ 0.3} \\
Granite4.1 8B & 82.3 $\pm$ 1.5 & 74.5 $\pm$ 0.1 & 46.9 $\pm$ 0.8 & 72.0 $\pm$ 1.2 & 85.8 $\pm$ 0.3 \\
Granite4.1 30B & 87.0 $\pm$ 0.1 & 74.0 $\pm$ 0.1 & 48.6 $\pm$ 0.9 & 78.1 $\pm$ 1.4 & 80.1 $\pm$ 0.3 \\
Ministral3 3B & 86.5 $\pm$ 0.4 & 73.4 $\pm$ 0.2 & 53.5 $\pm$ 1.0 & 78.2 $\pm$ 1.4 & 84.2 $\pm$ 0.3 \\
Mistral-Nemo 12B & 92.1 $\pm$ 0.6 & 75.0 $\pm$ 0.1 & 59.4 $\pm$ 1.2 & 86.2 $\pm$ 1.4 & 85.8 $\pm$ 0.3 \\
Mistral-Small 3.2 24B & 77.4 $\pm$ 0.9 & 74.2 $\pm$ 0.1 & 58.8 $\pm$ 1.4 & 69.9 $\pm$ 0.7 & 84.2 $\pm$ 0.3 \\
Llama3.1 8B & 88.4 $\pm$ 1.1 & 74.8 $\pm$ 0.2 & \cellcolor{aliceblue}\textbf{69.8 $\pm$ 1.4} & 82.0 $\pm$ 1.0 & 85.6 $\pm$ 0.3 \\
Llama3.3 70B & 78.2 $\pm$ 1.4 & 73.2 $\pm$ 0.1 & 65.7 $\pm$ 1.5 & 79.3 $\pm$ 1.3 & 83.7 $\pm$ 0.3 \\
Qwen3.5 9B & 79.2 $\pm$ 1.4 & 74.0 $\pm$ 0.2 & 57.9 $\pm$ 1.1 & 78.8 $\pm$ 1.3 & 85.5 $\pm$ 0.4 \\
Qwen3.5 35B & 80.5 $\pm$ 0.7 & 72.2 $\pm$ 0.2 & 64.4 $\pm$ 1.4 & 74.7 $\pm$ 1.3 & 83.8 $\pm$ 0.4 \\
\midrule
\multicolumn{6}{c}{\textit{Reports normalised utility $\val$~$\uparrow$}} \\
\midrule
Granite4.1 3B & \cellcolor{aliceblue}\textbf{87.5 $\pm$ 0.3} & 50.6 $\pm$ 0.3 & 26.0 $\pm$ 1.2 & 74.3 $\pm$ 2.9 & \cellcolor{aliceblue}\textbf{74.3 $\pm$ 0.6} \\
Granite4.1 8B & 69.6 $\pm$ 3.0 & 50.5 $\pm$ 0.3 & 7.8 $\pm$ 1.5 & 65.6 $\pm$ 2.2 & 73.7 $\pm$ 0.6 \\
Granite4.1 30B & 87.3 $\pm$ 0.3 & 50.4 $\pm$ 0.2 & 13.2 $\pm$ 1.5 & 72.9 $\pm$ 2.9 & 69.4 $\pm$ 0.5 \\
Ministral3 3B & 86.4 $\pm$ 0.7 & 48.5 $\pm$ 0.5 & 18.4 $\pm$ 2.0 & 73.0 $\pm$ 2.8 & 72.0 $\pm$ 0.6 \\
Mistral-Nemo 12B & 84.2 $\pm$ 1.2 & 50.3 $\pm$ 0.2 & 20.0 $\pm$ 2.3 & 72.7 $\pm$ 2.8 & 71.9 $\pm$ 0.6 \\
Mistral-Small 3.2 24B & 82.2 $\pm$ 1.7 & 49.7 $\pm$ 0.3 & 35.2 $\pm$ 2.8 & 56.1 $\pm$ 1.2 & 72.5 $\pm$ 0.6 \\
Llama3.1 8B & 77.1 $\pm$ 2.3 & 49.8 $\pm$ 0.4 & 40.4 $\pm$ 2.7 & 64.5 $\pm$ 2.0 & 71.5 $\pm$ 0.7 \\
Llama3.3 70B & 69.9 $\pm$ 2.9 & 50.4 $\pm$ 0.2 & \cellcolor{aliceblue}\textbf{48.9 $\pm$ 2.9} & 75.7 $\pm$ 2.6 & 73.8 $\pm$ 0.6 \\
Qwen3.5 9B & 72.0 $\pm$ 2.8 & 49.6 $\pm$ 0.4 & 23.2 $\pm$ 2.3 & 74.7 $\pm$ 2.6 & 73.0 $\pm$ 0.6 \\
Qwen3.5 35B & 83.7 $\pm$ 1.4 & 50.3 $\pm$ 0.3 & 42.0 $\pm$ 2.8 & \cellcolor{aliceblue}\textbf{77.2 $\pm$ 2.6} & 71.4 $\pm$ 0.7 \\
\midrule
\multicolumn{6}{c}{\textit{Reports normalised cost $\cost$~$\downarrow$}} \\
\midrule
Granite4.1 3B & \cellcolor{aliceblue}\textbf{0.0 $\pm$ 0.0} & 0.1 $\pm$ 0.0 & \cellcolor{aliceblue}\textbf{0.0 $\pm$ 0.0} & \cellcolor{aliceblue}\textbf{0.0 $\pm$ 0.0} & \cellcolor{aliceblue}\textbf{0.0 $\pm$ 0.0} \\
Granite4.1 8B & 5.0 $\pm$ 0.0 & 1.5 $\pm$ 0.0 & 13.9 $\pm$ 0.3 & 21.6 $\pm$ 1.1 & 2.0 $\pm$ 0.1 \\
Granite4.1 30B & 13.3 $\pm$ 0.0 & 2.4 $\pm$ 0.1 & 16.1 $\pm$ 0.9 & 16.7 $\pm$ 0.0 & 9.3 $\pm$ 0.4 \\
Ministral3 3B & 13.4 $\pm$ 0.2 & 1.7 $\pm$ 0.0 & 11.4 $\pm$ 0.2 & 16.6 $\pm$ 0.2 & 3.6 $\pm$ 0.1 \\
Mistral-Nemo 12B & 0.1 $\pm$ 0.0 & 0.2 $\pm$ 0.0 & 1.2 $\pm$ 0.1 & 0.4 $\pm$ 0.1 & 0.3 $\pm$ 0.0 \\
Mistral-Small 3.2 24B & 27.4 $\pm$ 0.8 & 1.3 $\pm$ 0.0 & 17.7 $\pm$ 0.7 & 16.3 $\pm$ 0.8 & 4.2 $\pm$ 0.1 \\
Llama3.1 8B & 0.2 $\pm$ 0.0 & 0.1 $\pm$ 0.0 & 0.9 $\pm$ 0.0 & 0.4 $\pm$ 0.0 & 0.3 $\pm$ 0.0 \\
Llama3.3 70B & 13.6 $\pm$ 0.0 & 3.9 $\pm$ 0.1 & 17.5 $\pm$ 0.6 & 17.0 $\pm$ 0.0 & 6.4 $\pm$ 0.2 \\
Qwen3.5 9B & 13.6 $\pm$ 0.0 & 1.5 $\pm$ 0.0 & 7.4 $\pm$ 0.0 & 17.2 $\pm$ 0.0 & 2.0 $\pm$ 0.0 \\
Qwen3.5 35B & 22.8 $\pm$ 0.0 & 6.0 $\pm$ 0.3 & 13.2 $\pm$ 0.0 & 27.8 $\pm$ 0.0 & 3.7 $\pm$ 0.2 \\
\bottomrule
\end{tabular}}
\label{tab:fixed_model_results_full}
\end{table}

\begin{figure}[H]
  \centering
  \includegraphics[width=1\textwidth]{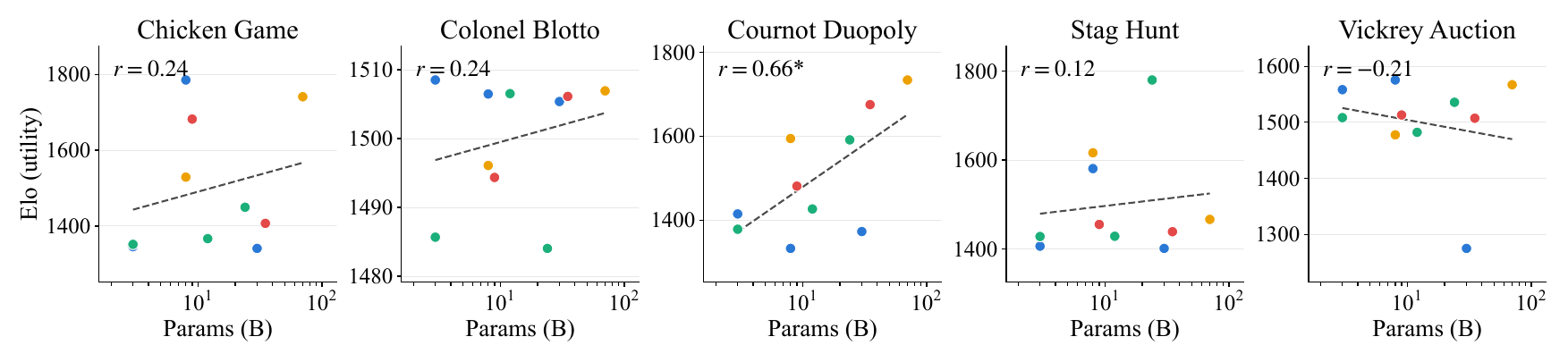} 
  \caption{Model elo score (bootstrapped) vs. scale. \textcolor{granitecolor}{\textbf{Granite}}, \textcolor{mistralcolor}{\textbf{Mistral}}, \textcolor{llamacolor}{\textbf{Llama}}, \textcolor{qwencolor}{\textbf{Qwen3.5}}.}
  \label{fig:detailed-scale}
\end{figure}

\subsection{Selector vs. Fixed Model}
\label{app:selector_vs_fixed}

\begin{table}[H]
\centering
\footnotesize
\setlength{\tabcolsep}{5pt}
\renewcommand{\arraystretch}{1.05}
\caption{Normalised net gain $\gain\uparrow$ (top) and in-game utility $\val\uparrow$ (bottom) acquired by each selector and fixed model from full pairwise matchups against each other, and across 5 games.
We show the mean $\pm$ 95\% CI, $\times100$, using $T=300$ rounds, $N=1$ seed per run, 
and \colorbox{aliceblue}{\textbf{highlight}} the best score for each game.
Selectors' learning signal was set as their per-round gain $g_{i, t}$.
}
\resizebox{\linewidth}{!}{%
\begin{tabular}{lrrrrr}
\toprule
\rowcolor{oldlace}\textbf{Method} & \textbf{Chicken Game} & \textbf{Colonel Blotto} & \textbf{Cournot Duopoly} & \textbf{Stag Hunt} & \textbf{Vickrey Auction} \\
\midrule
\multicolumn{6}{c}{\textit{Reports normalised gain $\gain$~$\uparrow$}} \\
\midrule
\textit{Per-Opponent Full-Information} & 92.8 $\pm$ 0.4 & 73.0 $\pm$ 0.2 & 80.7 $\pm$ 1.5 & 92.2 $\pm$ 1.0 & 86.7 $\pm$ 0.5 \\
\midrule
Granite4.1 3B & 89.7 $\pm$ 0.1 & 71.6 $\pm$ 0.1 & 64.0 $\pm$ 0.5 & 89.2 $\pm$ 0.8 & 83.4 $\pm$ 0.2 \\
Granite4.1 8B & 84.5 $\pm$ 0.8 & 70.5 $\pm$ 0.1 & 49.8 $\pm$ 0.5 & 81.1 $\pm$ 0.7 & 82.0 $\pm$ 0.2 \\
Granite4.1 30B & 77.3 $\pm$ 0.1 & 59.1 $\pm$ 0.1 & 40.9 $\pm$ 0.5 & 76.6 $\pm$ 0.8 & 68.6 $\pm$ 0.2 \\
Ministral3 3B & 86.5 $\pm$ 0.2 & 67.3 $\pm$ 0.2 & 53.3 $\pm$ 0.7 & 83.6 $\pm$ 0.8 & 78.9 $\pm$ 0.2 \\
Mistral-Nemo 12B & \cellcolor{aliceblue}\textbf{91.9 $\pm$ 0.4} & \cellcolor{aliceblue}\textbf{75.1 $\pm$ 0.1} & 61.0 $\pm$ 0.8 & \cellcolor{aliceblue}\textbf{90.9 $\pm$ 0.8} & \cellcolor{aliceblue}\textbf{85.8 $\pm$ 0.2} \\
Mistral-Small 3.2 24B & 80.7 $\pm$ 0.6 & 64.2 $\pm$ 0.1 & 57.2 $\pm$ 0.9 & 66.6 $\pm$ 0.4 & 75.7 $\pm$ 0.2 \\
Llama3.1 8B & 90.4 $\pm$ 0.7 & 74.6 $\pm$ 0.1 & \cellcolor{aliceblue}\textbf{72.6 $\pm$ 0.9} & 81.9 $\pm$ 0.7 & 85.6 $\pm$ 0.3 \\
Llama3.3 70B & 71.9 $\pm$ 0.9 & 58.2 $\pm$ 0.1 & 61.3 $\pm$ 0.9 & 73.8 $\pm$ 0.8 & 69.8 $\pm$ 0.2 \\
Qwen3.5 9B & 80.4 $\pm$ 0.8 & 65.7 $\pm$ 0.1 & 60.3 $\pm$ 0.8 & 81.2 $\pm$ 0.8 & 77.0 $\pm$ 0.2 \\
Qwen3.5 35B & 42.5 $\pm$ 0.5 & 25.1 $\pm$ 0.1 & 25.1 $\pm$ 0.9 & 42.4 $\pm$ 0.8 & 35.6 $\pm$ 0.3 \\
\midrule
Thompson Sampler (TS) & 85.1 $\pm$ 0.5 & 71.2 $\pm$ 0.2 & 47.7 $\pm$ 0.8 & 76.1 $\pm$ 0.8 & 74.8 $\pm$ 0.1 \\
RouterDC & 84.3 $\pm$ 0.1 & 68.6 $\pm$ 0.1 & 36.2 $\pm$ 0.1 & 74.4 $\pm$ 0.8 & 72.8 $\pm$ 0.2 \\
Causal LLM Router & 78.6 $\pm$ 0.7 & 63.2 $\pm$ 0.5 & 31.3 $\pm$ 0.7 & 74.0 $\pm$ 0.9 & 64.8 $\pm$ 0.5 \\
LLM Selector & 61.0 $\pm$ 0.6 & 46.9 $\pm$ 0.3 & 26.9 $\pm$ 0.5 & 46.6 $\pm$ 1.2 & 54.4 $\pm$ 0.2 \\
\midrule
\multicolumn{6}{c}{\textit{Reports normalised utility $\val$~$\uparrow$}} \\
\midrule
\textit{Per-Opponent Full-Information} & 88.9 $\pm$ 0.7 & 50.6 $\pm$ 0.3 & 43.0 $\pm$ 2.6 & 88.1 $\pm$ 2.1 & 75.1 $\pm$ 0.7 \\
\midrule
Granite4.1 3B & 86.7 $\pm$ 0.2 & 50.5 $\pm$ 0.1 & 35.3 $\pm$ 1.0 & 85.6 $\pm$ 1.5 & \cellcolor{aliceblue}\textbf{74.1 $\pm$ 0.5} \\
Granite4.1 8B & 78.2 $\pm$ 1.6 & 50.3 $\pm$ 0.1 & 8.8 $\pm$ 1.0 & 71.3 $\pm$ 1.4 & 73.2 $\pm$ 0.5 \\
Granite4.1 30B & 86.7 $\pm$ 0.2 & 50.3 $\pm$ 0.1 & 13.9 $\pm$ 1.0 & 85.4 $\pm$ 1.5 & 69.4 $\pm$ 0.4 \\
Ministral3 3B & 86.8 $\pm$ 0.5 & 48.5 $\pm$ 0.3 & 20.3 $\pm$ 1.4 & 81.1 $\pm$ 1.6 & 71.6 $\pm$ 0.5 \\
Mistral-Nemo 12B & 83.7 $\pm$ 0.8 & 50.2 $\pm$ 0.2 & 22.1 $\pm$ 1.5 & 81.7 $\pm$ 1.7 & 71.6 $\pm$ 0.5 \\
Mistral-Small 3.2 24B & 82.0 $\pm$ 1.2 & 49.0 $\pm$ 0.2 & 35.0 $\pm$ 1.8 & 53.9 $\pm$ 0.8 & 72.1 $\pm$ 0.5 \\
Llama3.1 8B & 80.8 $\pm$ 1.4 & 49.3 $\pm$ 0.3 & 45.2 $\pm$ 1.7 & 63.8 $\pm$ 1.4 & 71.2 $\pm$ 0.5 \\
Llama3.3 70B & 77.7 $\pm$ 1.8 & 50.4 $\pm$ 0.1 & \cellcolor{aliceblue}\textbf{56.5 $\pm$ 1.8} & 81.5 $\pm$ 1.6 & 73.6 $\pm$ 0.5 \\
Qwen3.5 9B & 79.1 $\pm$ 1.6 & 49.8 $\pm$ 0.2 & 39.0 $\pm$ 1.7 & 80.7 $\pm$ 1.6 & 72.4 $\pm$ 0.5 \\
Qwen3.5 35B & 85.0 $\pm$ 0.9 & 50.2 $\pm$ 0.2 & 50.2 $\pm$ 1.8 & 84.8 $\pm$ 1.6 & 71.2 $\pm$ 0.6 \\
\midrule
Thompson Sampler (TS) & 87.4 $\pm$ 0.7 & 50.4 $\pm$ 0.1 & 39.4 $\pm$ 1.4 & 78.0 $\pm$ 1.3 & 73.4 $\pm$ 0.3 \\
RouterDC & \cellcolor{aliceblue}\textbf{87.8 $\pm$ 0.2} & 50.5 $\pm$ 0.1 & 19.6 $\pm$ 0.3 & 76.1 $\pm$ 1.5 & 72.8 $\pm$ 0.3 \\
Causal LLM Router & 79.7 $\pm$ 1.1 & 50.1 $\pm$ 0.2 & 27.1 $\pm$ 1.3 & 72.4 $\pm$ 1.4 & 72.2 $\pm$ 0.4 \\
LLM Selector & 80.6 $\pm$ 1.2 & 50.1 $\pm$ 0.2 & 18.1 $\pm$ 1.0 & \cellcolor{aliceblue}\textbf{87.0 $\pm$ 2.3} & 71.5 $\pm$ 0.5 \\
\bottomrule
\end{tabular}}
\label{tab:selectors_vs_fixed_full}
\end{table}